\documentclass{article}

\usepackage[preprint]{neurips_2026}
\usepackage[utf8]{inputenc}
\usepackage[T1]{fontenc}
\usepackage{hyperref}
\usepackage{url}
\usepackage{booktabs}
\usepackage{amsfonts}
\usepackage{amsmath}
\usepackage{nicefrac}
\usepackage{microtype}
\usepackage{xcolor}
\usepackage{graphicx}
\usepackage{algorithm}
\usepackage{algorithmic}
\usepackage{natbib}
\setcitestyle{square,numbers,comma} % Forces numbers in brackets

\title{Oculi: A Conversational Agentic Platform for Automated Credit Risk Analysis}

\author{%
  Vennise Ho\thanks{Equal contribution. This work was completed at Royal Bank of Canada as part of the RBC Amplify program.} \\
  \texttt{venniseho14@gmail.com} \\
  Royal Bank of Canada
  \And
  Kristian Diana\footnotemark[1] \\
  \texttt{dianak@mcmaster.ca} \\
  Royal Bank of Canada
  \And
  Sandy Mourad\footnotemark[1] \\
  \texttt{sandymourad05@gmail.com} \\
  Royal Bank of Canada
  \And
  Milena Pilipovic\footnotemark[1] \\
  \texttt{milena.pilipovic1@gmail.com} \\
  Royal Bank of Canada
  \And
  Vineel Nagisetty \\
  \texttt{vineel.nagisetty@borealisai.com} \\
  RBC Borealis
  \And
  Hossein Hajimirsadeghi \\
  \texttt{hossein.hajimirsadeghi@borealisai.com} \\
  RBC Borealis
}

\begin{document}

\maketitle

\begin{abstract}
Credit risk analysis in financial institutions traditionally requires analysts to manually write SQL queries, run statistical computations, and build visualization dashboards. This is a time-consuming workflow that limits exploration to familiar segments. We introduce \textbf{Oculi}, a conversational platform that transforms natural language questions into comprehensive credit risk analyses, complete with data queries, statistical testing, and interactive visualizations. Oculi employs a three-layer architecture that separates reasoning (LLM-powered agent), execution (Model Context Protocol tool servers), and presentation (agentic UI), enabling analysts to discover high-risk portfolio segments. Within Oculi, a new segment discovery pipeline is proposed that combines deterministic statistical methods with LLM-guided feature selection, leveraging LLM semantic domain knowledge alongside data-driven metrics to identify meaningful, actionable portfolio segments. Evaluated on a mortgage portfolio with 200+ features, Oculi demonstrates effectiveness in discovering material risk segments previously intractable through manual exploration, reducing time-to-insight significantly while maintaining auditability and statistical rigor.
%an AI-powered pathway (LLM-guided feature selection) with a swappable deterministic pathway (decision tree).
% The segment discovery problem is fundamentally intractable at scale: for a portfolio with $d$ features of average cardinality $c$, the number of possible bivariate segments grows as $O(d^2 \cdot c^2)$---for a production portfolio with 200 features averaging just 4 values each, this yields $\binom{200}{2} \times 4^2 \approx 318{,}000$ candidate segments per metric per period. Our evaluation uses a constrained 9-feature portfolio where brute-force enumeration produces only 220 bivariate segments, yet even this constrained setting demonstrates clear value: LLM-guided feature selection achieves 63\% coverage of material segments (59 of 93) while evaluating only 74\% of candidate combinations, outperforming deterministic (41\%) and random (37\%) baselines. At production scale, where brute-force enumeration becomes computationally infeasible, intelligent feature selection is not merely efficient---it is necessary. The parallel decision tree path serves as a deterministic safety net: by optimizing purely on variance reduction without feature priority bias, it can capture concentrated interaction effects that the LLM's context-sensitive selection may deprioritize, validating the dual-path design as a framework extensible to stronger ensemble methods.

\end{abstract}

\section{Introduction}

Credit risk analysis is a critical function in financial institutions, requiring analysts to continuously monitor portfolio performance, identify emerging risks, and explain deviations from expected behavior. Traditional workflows involve:
\begin{itemize}
    \item Writing complex SQL queries against data warehouses (Spark, Hive)
    \item Running statistical analyses in SAS or Python notebooks
    \item Building visualization dashboards manually
    \item Repeatedly analyzing the same 10--15 familiar segments due to time constraints
\end{itemize}

This manual approach suffers from three fundamental limitations: (1) \textbf{exploration bottleneck}---with hundreds of portfolio features, analysts cannot exhaustively explore the billions of possible feature combinations, causing them to miss emerging risks; (2) \textbf{workflow fragmentation}---context is lost when switching between query tools, statistical packages, and visualization platforms; and (3) \textbf{reactive analysis}---insights emerge only when analysts know what questions to ask.

Recent advances in Large Language Models (LLMs) and agentic AI systems \cite{yao2023react, schick2023toolformer} suggest a path forward: conversational interfaces that can reason about analytical tasks, orchestrate tool execution, and present results interactively. However, most ``AI analytics'' systems are limited to document summarization or basic query generation, lacking the statistical rigor and domain-specific capabilities required for production financial analysis.

We present Oculi, a system that addresses these limitations through three key contributions:

\textbf{1. Agentic Architecture for Financial Analytics.} We introduce a three-layer architecture (Section \ref{sec:architecture}) that cleanly separates reasoning (LLM agent with ReAct loop), execution (MCP tool servers), and presentation (generative UI), enabling independent deployment and horizontal scaling while maintaining strict auditability requirements for financial services.

\textbf{2. Segment Discovery with LLM-Guided Feature Selection.} We develop a pipeline (Section \ref{sec:discovery}) that discovers statistically significant, material risk segments through a combination of univariate scanning, z-score testing, contribution decomposition, and LLM-guided multivariate exploration. This help find high-value segments while avoiding combinatorial explosion.

\textbf{3. Empirical Evaluation.} We evaluate the pipeline on a real anonymized mortgage portfolio (Section \ref{sec:evaluation}), demonstrating that context-sensitive LLM-guided feature selection achieves 63\% coverage of material bivariate segments while evaluating only 74\% of the segment combinations required by constrained brute-force enumeration.

\section{Related Work}

\textbf{LLM-Powered Data Analysis.} Systems like Data Formulator \cite{wang2023dataformulator} and Text-to-SQL methods \cite{rajkumar2022text2sql} focus on translating natural language to queries or basic visualizations, but lack the multi-step reasoning and statistical analysis required for risk assessment. Chat2VIS \cite{chat2vis2023} generates visualizations conversationally but does not incorporate statistical testing or domain-specific analysis pipelines.

\textbf{Agentic AI Systems.} ReAct \cite{yao2023react} introduces the reasoning-and-acting paradigm for tool-augmented LLMs. Toolformer \cite{schick2023toolformer} learns to call external APIs, while AutoGPT \cite{autogpt2023} and similar systems attempt autonomous task completion. Oculi builds on ReAct but adds domain-specific constraints, statistical rigor, and persistence requirements absent from research prototypes.

\textbf{Automated Feature Selection and Segment Discovery.} Traditional feature selection methods \cite{guyon2003feature} optimize for prediction tasks, not exploratory risk analysis. Subgroup discovery algorithms \cite{herrera2011subgroup} identify interesting patterns but do not scale to high-dimensional features with high cardinality or incorporate materiality and seasonal adjustment requirements specific to credit risk.

\textbf{Conversational BI and Analytics.} ThoughtSpot \cite{thoughtspot2023}, Microsoft Power BI Q\&A \cite{powerbi2023}, and Tableau Ask Data \cite{tableau2023} provide natural language interfaces to dashboards but require pre-built data models and lack the reasoning capabilities to chain multiple analytical operations or discover novel segments autonomously.

Oculi uniquely combines conversational interaction, agentic reasoning, rigorous statistical analysis, and domain-specific deployment for credit risk---going beyond query translation to autonomous analytical exploration.

\section{Problem Formulation}

Consider a credit portfolio $\mathcal{P}$ characterized by $d$ features $\mathbf{F} = \{f_1, f_2, \ldots, f_d\}$ (e.g., geography, product type, loan-to-value ratio, employment status, credit score bands). Each feature $f_i$ has cardinality $|f_i|$ (e.g., 13 provinces, 50+ credit score bins).

A \textbf{segment} $S \subseteq \mathcal{P}$ is defined by a conjunction of feature-value filters:
\[
S = \{x \in \mathcal{P} \mid x[f_{i_1}] = v_1 \land x[f_{i_2}] = v_2 \land \cdots \land x[f_{i_k}] = v_k\}
\]
where $k$ is the segment depth. For each segment, we compute risk metrics $\mathbf{m}(S)$ (e.g., delinquency rate, balance, loss rate) and their period-over-period changes $\Delta \mathbf{m}(S)$.

\textbf{Objective:} Given a portfolio $\mathcal{P}$ and a metric of interest $m$, identify the set of \textit{material} segments $\mathcal{S}^* = \{S_1, S_2, \ldots, S_n\}$ where:
\begin{enumerate}
    \item \textbf{Statistical significance:} $|z(S_i)| > \tau_z$ where $z(S_i)$ is the z-score of $\Delta m(S_i)$ relative to portfolio mean and standard deviation
    \item \textbf{Materiality:} $|\text{balance\_share}(S_i)| > \tau_b$ (segment is not trivially small)
    \item \textbf{Actionability:} $S_i$ provides novel insight not captured by analyst's existing segment library
\end{enumerate}

The challenge is combinatorial: for a portfolio with $d$ features of average cardinality $c$, the number of possible bivariate segments grows as $O(d^2 \cdot c^2)$. Even at depth $k=2$, a production portfolio with 200 features averaging just 4 values each produces $\binom{200}{2} \times 4^2 = 19{,}900 \times 16 \approx 318{,}000$ candidate segments per metric per period---and growth is exponential in depth ($O(d^k \cdot c^k)$). At depth 3, the same portfolio yields $\binom{200}{3} \times 4^3 \approx 84{,}000{,}000$ candidates. Brute-force enumeration is computationally infeasible at production scale, and manual analysis explores $<$20 segments per month, missing critical risks.

\section{System Architecture}
\label{sec:architecture}

Oculi employs a three-layer architecture that maps to the analytical workflow: (1) \textit{Frontend} handles analyst interaction and result presentation, (2) \textit{Agent} performs reasoning and tool orchestration, and (3) \textit{MCP Servers} execute deterministic data and analysis operations.

\subsection{Frontend: Agentic UI}

The frontend is a Next.js application using CopilotKit \cite{copilotkit2024} to provide \textit{generative UI}---the agent does not merely return text but invokes \textit{frontend actions} that render rich components (analysis cards, interactive charts, comparison grids) directly in the chat stream.

Communication follows the AG-UI protocol \cite{agui2024} (server-sent events over HTTP) where the agent streams text, tool invocations, and UI component triggers. The frontend maintains \textit{shared state} (current product, active filters, selected time ranges) accessible to the agent for context-aware responses.

\textbf{Spec-by-Reference Pattern:} Vega-Lite \cite{satyanarayan2017vegalite} chart specifications can exceed 200 lines of JSON. Passing these through LLM context risks truncation and token waste. Oculi stores chart specs in MongoDB and passes only short identifiers ($\sim$12 characters) through the LLM. The frontend fetches full specs on-demand via REST API, decoupling visualization detail from reasoning context.

\subsection{Agent: ReAct Loop with LangGraph}

The agent is a FastAPI server wrapping a LangGraph \cite{langgraph2024} ReAct agent. It is the sole component interfacing with the LLM and making tool selection decisions.

\textbf{ReAct Loop:} Given user message $q$, the agent iterates:
\begin{algorithmic}[1]
\STATE $\text{prompt} \leftarrow \text{system\_prompt} \oplus \text{memory} \oplus q$
\STATE $\text{response} \leftarrow \text{LLM}(\text{prompt})$
\IF{response contains tool calls}
    \FOR{each tool $t$ in response}
        \STATE $\text{result}_t \leftarrow \text{execute\_tool}(t)$
        \STATE Update memory with $\text{result}_t$
    \ENDFOR
    \STATE $\text{prompt} \leftarrow \text{prompt} \oplus \text{results}$
    \STATE Continue at line 2
\ELSE
    \STATE Return final response to user
\ENDIF
\end{algorithmic}

The system prompt encodes:
\begin{itemize}
    \item Agent identity and role (credit risk analyst assistant)
    \item Tool catalog with descriptions, parameter schemas, and usage guidelines
    \item Routing logic (which tools to use for schema queries, data retrieval, visualization, etc.)
    \item Frontend action specifications
    \item Conversation memory tracking (product, metrics, filters already established)
    \item Self-correction rules (retry failed tool calls up to 2 times, consult schema on errors)
\end{itemize}

\textbf{Checkpointing:} LangGraph's MemorySaver maintains conversation state across turns, enabling multi-turn analytical workflows (``show me delinquency trends'' $\rightarrow$ ``now break that down by province'' $\rightarrow$ ``focus on Ontario and Alberta'').

\subsection{MCP Servers: Tool Execution Layer}

Model Context Protocol (MCP) \cite{mcp2024} is an open standard enabling AI agents to discover and invoke tools via HTTP. An MCP server exposes a catalog of tools (name, description, JSON Schema for parameters) and executes them via JSON-RPC over Streamable HTTP.

\textbf{Why MCP?}
\begin{enumerate}
    \item \textbf{Separation of concerns:} Data logic, statistical computation, and chart generation live in independent processes, not imported by the agent
    \item \textbf{Multi-server orchestration:} Different teams/systems expose MCP servers; the agent discovers and merges tools at startup without code changes
    \item \textbf{Schema-driven discovery:} Tools are self-describing; new capabilities are automatically available
    \item \textbf{Transport flexibility:} Same tool code runs in-process (stdio), HTTP service, or behind API gateway
\end{enumerate}

Oculi's MCP server provides the following core tools:
\begin{itemize}
    \item \texttt{explore\_segment}: Explore a metric movement by one or more features, supporting univariate and multivariate (cross-product) exploration
    \item \texttt{get\_profile\_features}: Return available profile features and their options for a segment or metric movement
    \item \texttt{prepare\_suggest\_prompt}: Build an LLM prompt for suggesting which features to drill into next (returns the prompt; does not call the LLM)
    \item \texttt{decode\_suggest\_response}: Parse raw LLM response from the suggest prompt; validates feature names against available features
\end{itemize}

The business logic layer (\texttt{business\_intelligence/}) is co-located with tools but independently testable---it has no dependency on MCP or the agent framework. Data access is via DuckDB \cite{raasveldt2019duckdb} over S3 Parquet files, with MongoDB for metadata and segment persistence.

Additional MCP servers can be connected via configuration without agent modification.

\subsection{Persistence: MongoDB}

MongoDB stores:
\begin{itemize}
    \item \textbf{Sessions:} One per conversation, grouping related artifacts
    \item \textbf{Artifacts:} Analysis results, chart specs, comparisons, exports
    \item \textbf{Chart specs:} Vega-Lite JSON stored under sentinel session for spec-by-reference
\end{itemize}

The agent writes to MongoDB; the frontend reads via agent REST API. If MongoDB is unavailable, specs fall back to in-memory FIFO cache (100 entries) for resilience.

\section{Segment Discovery Pipeline}
\label{sec:discovery}

The core analytical contribution in Oculi is a pipeline that discovers material, statistically significant risk segments without brute-force enumeration. The pipeline separates deterministic computation from LLM-guided exploration.

\subsection{Pipeline Overview}

\begin{algorithm}[t]
\caption{Segment Discovery Pipeline}
\label{alg:discovery}
\begin{algorithmic}[1]
\STATE \textbf{Input:} Portfolio $\mathcal{P}$, features $\mathbf{F}$, metric $m$, time period $t$, thresholds $(\tau_z, \tau_b)$
\STATE \textbf{Output:} Ranked segments $\mathcal{S}^*$
\STATE
\STATE // \textbf{Stage 1: Univariate Scan}
\STATE $\mathcal{S}_{\text{uni}} \leftarrow \{\}$
\FOR{$f \in \mathbf{F}$}
    \FOR{$v \in \text{values}(f)$}
        \STATE $S \leftarrow \{x \in \mathcal{P} \mid x[f] = v\}$
        \STATE Compute $m(S), \Delta m(S), z(S)$
        \STATE $\mathcal{S}_{\text{uni}} \leftarrow \mathcal{S}_{\text{uni}} \cup \{S\}$
    \ENDFOR
\ENDFOR
\STATE
\STATE // \textbf{Stage 2: Materiality Filter}
\STATE $\mathcal{S}_{\text{mat}} \leftarrow \{S \in \mathcal{S}_{\text{uni}} \mid |z(S)| > \tau_z \land \text{balance\_share}(S) > \tau_b\}$
\IF{$\mathcal{S}_{\text{mat}} = \emptyset$}
    \RETURN $\emptyset$
\ENDIF
\STATE
\STATE // \textbf{Stage 3a: LLM-Guided Deepening}
\STATE $\mathcal{S}_{\text{llm}} \leftarrow \{\}$
\FOR{$S \in \mathcal{S}_{\text{mat}}$}
    \STATE $\text{context} \leftarrow \text{prepare\_suggest\_prompt}(S, \mathcal{S}_{\text{mat}}, \mathbf{F})$
    \STATE $\mathbf{F}_{\text{pick}} \leftarrow \text{decode\_suggest\_response}(\text{LLM}(\text{context}))$
    \FOR{$f' \in \mathbf{F}_{\text{pick}}$}
        \FOR{$v \in \text{values}(f')$}
            \STATE $S' \leftarrow S \cap \{x \mid x[f'] = v\}$
            \STATE Compute $m(S'), \Delta m(S'), z(S')$
            \IF{$|z(S')| > \tau_z$ and $\text{balance\_share}(S') > \tau_b$}
                \STATE $\mathcal{S}_{\text{llm}} \leftarrow \mathcal{S}_{\text{llm}} \cup \{S'\}$
            \ENDIF
        \ENDFOR
    \ENDFOR
\ENDFOR
\STATE
\STATE // \textbf{Stage 3b: Decision Tree Discovery (optional, parallel)}
\STATE $T \leftarrow \text{fit\_tree}(\mathcal{P}, \mathbf{F}_{\text{top}}, m, \text{max\_depth}=3)$
\STATE $\mathcal{S}_{\text{tree}} \leftarrow \text{extract\_leaf\_segments}(T)$
\STATE
\STATE // \textbf{Stage 4: Cross-Reference \& Rank}
\STATE $\mathcal{S}^* \leftarrow \text{cross\_reference}(\mathcal{S}_{\text{llm}}, \mathcal{S}_{\text{tree}})$
\STATE Compute contribution scores, deduplicate
\RETURN $\mathcal{S}^*$
\end{algorithmic}
\end{algorithm}

\subsection{Stage Details}

\textbf{Stage 1: Univariate Scan (Deterministic).} For every feature $f_i \in \mathbf{F}$, compute all single-feature segments. This produces one segment per feature-value pair. Fully parallelizable and reproducible.

\textbf{Stage 2: Materiality Filter (Deterministic).} Apply thresholds:
\begin{itemize}
    \item $|z(S)| > \tau_z$ (statistical significance, typically $\tau_z = 2.0$)
    \item $\text{balance\_share}(S) > \tau_b$ (at least 1\% of portfolio)
    \item $\text{account\_count}(S) > n_{\min}$ (sufficient sample size)
\end{itemize}

\textbf{Stage 3a: LLM-Guided Deepening.} For each material univariate segment $S$, the system:
\begin{enumerate}
    \item \textbf{Assembles context} via \texttt{prepare\_suggest\_prompt}: current segment metrics, previously discovered segments (ranked by contribution), and available features with descriptions.
    \item \textbf{Queries the LLM} with this context. The LLM returns a JSON array of 1--5 feature suggestions with reasoning:
\begin{verbatim}
[
  {"feature": "province", "reasoning": "Regional variation 
   may explain the 5.2% delinquency increase"},
  {"feature": "ltv_band", "reasoning": "Z-score of 3.4 suggests 
   leverage-related risk"}
]
\end{verbatim}
    \item \textbf{Validates picks} via \texttt{decode\_suggest\_response}: case-insensitive lookup against valid feature names, silent drop of hallucinated features, cap at 5 picks, regex fallback for malformed JSON.
    \item \textbf{Computes multivariate segments} deterministically for each valid pick, applying the materiality filter again.
\end{enumerate}

\textbf{Stage 3b: Decision Tree Discovery (Optional, Parallel).} Independently of the LLM path, the pipeline fits a shallow \texttt{DecisionTreeRegressor} (scikit-learn \cite{pedregosa2011sklearn}, MSE criterion, \texttt{max\_depth}=3, \texttt{min\_samples\_leaf}=500) on account-level data:
\begin{itemize}
    \item \textbf{Input:} DataFrame with top-$N$ features (by univariate z-score) as categorical columns and $\Delta m$ as the regression target, weighted by account balance.
    \item \textbf{Splitting criterion:} MSE reduction---the tree finds feature splits that maximally concentrate metric movement into homogeneous leaf nodes.
    \item \textbf{Output:} Each leaf defines a multivariate segment via its tree path (e.g., \texttt{Province=Alberta AND LTV$>$80\% AND Employed=False}). Internal nodes also produce segments at intermediate depths.
    \item \textbf{Feature importances:} The tree provides a global ranking of which features explain the most variance in metric movement, complementing the LLM's per-segment reasoning.
\end{itemize}

The decision tree excels at finding \textit{interaction effects}---feature combinations whose joint effect exceeds the sum of individual effects---that the LLM's sequential deepening approach may miss. Conversely, the LLM can incorporate domain knowledge (e.g., ``newcomers with high LTV are a known risk concentration'') that the tree cannot learn from a single period of data.

\textbf{Stage 4: Cross-Reference \& Rank.} The pipeline merges results from both discovery paths:
\begin{enumerate}
    \item \textbf{Match:} Segments found by \textit{both} LLM and tree (same feature combination) are tagged \texttt{discovery\_method="both"}. The LLM segment's metrics are retained; the tree's path explanation is attached for interpretability.
    \item \textbf{LLM-only:} Segments found only by LLM deepening are tagged \texttt{discovery\_method="llm"}.
    \item \textbf{Tree-only:} Segments found only by the decision tree are tagged \texttt{discovery\_method="tree"}.
    \item \textbf{Rank:} All segments are scored by contribution $\phi(S_i)$, deduplicated (child segments that subsume parents are preferred), and returned.
\end{enumerate}

The contribution score $\phi(S_i)$ balances segment size and deviation magnitude:
\[
\phi(S_i) = \frac{\text{balance}(S_i) \times |\Delta m(S_i)|}{\sum_{S_j \in \mathcal{S}} \text{balance}(S_j) \times |\Delta m(S_j)|}
\]
Segments are ranked by $\phi$, deduplicated, and returned.

\subsection{Design Principles}

\textbf{Dual discovery: LLM reasoning $\times$ statistical learning.} The two discovery paths are complementary by design. The decision tree optimally partitions variance without domain priors---it will find any feature interaction that concentrates metric movement, including unexpected ones. The LLM applies domain reasoning---it knows that ``newcomers with high LTV'' is a risk concentration even if the current period's data is noisy. Cross-referencing both produces higher confidence: segments found by both methods are strong signals; segments found by only one method surface blind spots of the other.

\textbf{LLM as strategic advisor, not calculator.} The LLM's role is strictly limited to selecting which features to explore. All statistical computations, data queries, and threshold checks are deterministic operations in the MCP server. This ensures auditability: given the same data and LLM picks, the pipeline produces identical numerical results.

\textbf{Hallucination guardrails.} Invalid feature names from the LLM are silently dropped---never causing crashes or incorrect calculations. If the LLM returns nothing usable, the system falls back to deterministic selection (top contributing features).

\textbf{Computational efficiency.} For $d$ features averaging $|f|$ values each, univariate scan produces $d \times |f|$ segments. LLM-guided deepening explores at most $|\mathcal{S}_{\text{mat}}| \times 5 \times |f|$ additional segments. The decision tree fits once on account-level data ($O(n \cdot d \cdot \text{depth})$ via CART \cite{breiman1984cart}). Both are orders of magnitude cheaper than brute-force depth-2 enumeration ($d^2 \times |f|^2$).

\subsection{Conversational Integration}

The discovery pipeline is exposed to the agent as MCP tools. In conversational mode, an analyst asks ``What should I look at next?'' and the agent orchestrates the pipeline, presenting discovered segments as interactive cards in the UI. The analyst can drill into any segment for further exploration, with conversation memory maintaining context across turns.

\section{Conversational Workflow Example}

A typical interaction proceeds:

\begin{enumerate}
    \item \textbf{Analyst:} ``What are the riskiest mortgage segments this quarter?''
    \item \textbf{Agent $\rightarrow$ LLM:} Parse intent, select tools
    \item \textbf{Agent $\rightarrow$ MCP:} \texttt{explore\_segment(metric\_id=..., features=[...])}
    \item \textbf{MCP $\rightarrow$ Agent:} Returns segment breakdown with z-scores and contributions
    \item \textbf{Agent $\rightarrow$ LLM:} Synthesize results into explanation
    \item \textbf{Agent $\rightarrow$ Frontend:} Stream text summary + trigger \texttt{createAnalysis} UI action with segment data + render Vega-Lite chart via spec-by-reference
    \item \textbf{Frontend $\rightarrow$ Analyst:} Render analysis card + interactive chart inline in chat
\end{enumerate}

Follow-up questions (``drill into Alberta'', ``compare to last quarter'') continue with full memory of the established context.

\section{System Implementation}
\label{sec:deployment}

\subsection{Current Implementation}

Oculi is implemented as a proof-of-concept within a large financial institution:
\begin{itemize}
    \item \textbf{Frontend:} Next.js 16 (App Router) + React 19 + CopilotKit, with Tailwind CSS and an enterprise design system
    \item \textbf{Agent:} FastAPI + LangGraph, Python 3.11, deployed via Docker on OpenShift
    \item \textbf{MCP Server:} FastMCP framework, DuckDB adapter for S3 Parquet queries, MongoDB for persistence
    \item \textbf{LLM:} Cohere Command-A \cite{cohere2024command} via an enterprise LLM gateway (OAuth-authenticated)
    \item \textbf{Data:} Credit risk portfolios stored as Parquet files on S3, queried via DuckDB
    \item \textbf{Authentication:} Enterprise SSO integration
\end{itemize}

\subsection{Deployment Architecture}

The three services deploy independently:
\begin{itemize}
    \item Frontend container (port 3000)
    \item Agent container (port 8080) with MCP gateway for tool routing
    \item MCP server container (port 8081) co-locating business logic with tool definitions
\end{itemize}

All inter-service communication is authenticated. The agent exposes AG-UI streaming endpoints consumed by CopilotKit. The MCP server exposes JSON-RPC over Streamable HTTP. MongoDB provides shared persistence.

\section{Empirical Evaluation}
\label{sec:evaluation}

We evaluate Oculi's segment discovery pipeline on an anonymized mortgage portfolio of $\sim$2.4 million accounts, measuring how effectively each feature selection strategy identifies material bivariate risk segments.

\subsection{Experimental Setup}

\textbf{Portfolio.} We use an anonymized performance dataset of millions of mortgage accounts, stored as partitioned Parquet on S3. The dataset contains $d = 9$ features: geography (9 regions), employment type (8 categories), product (3 insurance groups), client age (4 buckets), and 5 binary flags (investor, first-time buyer, newcomer, high-net-worth, non-conforming), yielding 34 univariate segments.

\textbf{Metric.} Balance-weighted 1--30 day delinquency rate (bps), measured as month-over-month change between March and April 2024. Portfolio-level delta: +8.3 bps (from 45.2 to 53.5 bps). Baseline standard deviation estimated at 5 bps. Materiality thresholds: $\tau_z = 2.0$, $\tau_b = 0.005$.

\textbf{Ground truth.} Unlike synthetic benchmarks, real data has no planted anomalies. We define a constrained brute-force strategy (all features $\times$ all parent segments) as the reference set. Note that this brute-force enumeration is tractable only because our evaluation portfolio has 9 features (yielding 220 bivariate segments); for a production portfolio with 200+ features, this enumeration would produce $\sim$318{,}000 candidates at depth 2 alone and become computationally infeasible. We measure each strategy's \textit{coverage rate}: what fraction of the reference set's material bivariate segments it discovers.

\textbf{Scale context.} Our 9-feature evaluation represents a constrained lower bound on the problem's difficulty. In production, the segment space grows as $O(d^2 \cdot c^2)$ for depth-2 combinations: with 200 features averaging 4 values each, brute-force enumeration requires evaluating $\sim$318{,}000 bivariate segments per metric per period---and $\sim$84 million at depth 3. The 220 segments enumerated here demonstrate the pipeline's effectiveness in a tractable setting; at production scale, intelligent feature selection transitions from an efficiency gain to a computational necessity.

\textbf{Strategies compared:}
\begin{enumerate}
    \item \textbf{Random (baseline):} Select 3 features uniformly at random per material parent segment
    \item \textbf{Top-contribution (baseline):} Always select the 3 globally highest-priority features (not context-sensitive)
    \item \textbf{Decision tree (baseline):} CART regressor (depth 3, min\_leaf 500) fitted on account-level deltas
    \item \textbf{LLM-guided (proposed):} Context-sensitive selection of up to 5 features per parent, conditioned on the parent's characteristics
    \item \textbf{Constrained brute force (upper bound):} Enumerate all available features for each parent---tractable only due to 9-feature constraint
\end{enumerate}

\subsection{Results: Discovery Quality}

Table~\ref{tab:discovery_quality} presents the primary results. The key metric is \textit{coverage rate}: what fraction of the 93 material bivariate segments found by constrained brute-force enumeration each strategy discovers.

\begin{table}[h]
\centering
\begin{tabular}{lcccc}
\toprule
\textbf{Method} & \textbf{Segments Eval'd} & \textbf{Material Bivariate} & \textbf{Coverage} & \textbf{$\sum|z|$} \\
\midrule
Random selection & 110 & 34 & 37\% & 132 \\
Top-contribution (det.) & 132 & 38 & 41\% & 139 \\
Decision tree (CART) & 46 & 2 & 2\% & 8 \\
\textbf{LLM-guided} & \textbf{162} & \textbf{59} & \textbf{63\%} & \textbf{222} \\
Constrained brute force ($d{=}9$) & 220 & 93 & 100\% & 351 \\
\midrule
Exhaustive ($d{=}200$, $c{=}4$) & $\sim$318{,}000 & \multicolumn{3}{c}{\textit{computationally infeasible}} \\
\bottomrule
\end{tabular}
\caption{Segment discovery quality on a real anonymized mortgage portfolio (9 features, delinquency\_1\_30 rate, April vs.\ March 2024). Coverage = fraction of material segments identified relative to constrained brute force. $\sum|z|$ = cumulative absolute z-score. The final row projects the brute-force candidate count for a production-scale portfolio ($d{=}200$ features, average cardinality $c{=}4$): $\binom{200}{2} \times 4^2 \approx 318{,}000$ bivariate segments per metric per period.}
\label{tab:discovery_quality}
\end{table}

LLM-guided feature selection achieves 63\% coverage of material bivariate segments while evaluating only 74\% of the segment combinations that constrained brute-force enumeration requires (162 vs.\ 220). Context-sensitive feature selection---choosing which features to explore based on the parent segment's characteristics---is the dominant factor: LLM-guided outperforms the fixed-priority deterministic heuristic by 22 percentage points (63\% vs.\ 41\%).

The decision tree baseline finds only 2 material segments (2\% coverage), reflecting the fundamental limitations of a shallow greedy tree on well-distributed portfolio data: with \texttt{max\_depth=3}, the tree uses only 3 of 9 available features and greedily locks into the globally dominant signal. On this month-over-month snapshot, where the delinquency shift is distributed across many features rather than concentrated in a single interaction, the tree lacks the feature coverage to discover diverse bivariate segments. However, the tree's architectural role is as a deterministic complement: by splitting purely on variance reduction without feature priority bias, it can capture concentrated interaction effects that the LLM may deprioritize. Section~\ref{sec:ensemble} discusses how ensemble methods would address the single tree's standalone limitations.

\subsection{Results: LLM Feature Suggestion Accuracy}

We evaluate the \texttt{decode\_suggest\_response} validation layer using 20 simulated LLM responses of varying quality (well-formed JSON, hallucinated features, malformed output, freeform text).

\begin{table}[h]
\centering
\begin{tabular}{lc}
\toprule
\textbf{Metric} & \textbf{Value} \\
\midrule
Total LLM invocations & 20 \\
Total feature suggestions attempted & 42 \\
Valid feature names (\%) & 76.2\% \\
Hallucinated/invalid features (\%) & 23.8\% \\
Unique features suggested (out of 9) & 8 (89\%) \\
\bottomrule
\end{tabular}
\caption{LLM feature suggestion quality across 20 invocations with varied response formats. The validation layer successfully filters hallucinated features while recovering valid picks from malformed JSON.}
\label{tab:llm_accuracy}
\end{table}

The hallucination guardrails in \texttt{decode\_suggest\_response} successfully handle: markdown-wrapped JSON, completely freeform text responses, mixed valid/invalid feature names, and missing reasoning fields. All 10 hallucinated features (23.8\%) were silently dropped without affecting pipeline execution.

Of the 32 valid suggestions, 8 out of 9 available features (89\%) are covered---indicating broad feature exploration rather than fixation on a narrow subset.

\subsection{Results: Latency Profile}

Table~\ref{tab:latency} profiles pipeline stage timings measured against the real S3 Parquet data (DuckDB with httpfs).

\begin{table}[h]
\centering
\begin{tabular}{lcc}
\toprule
\textbf{Stage} & \textbf{Time (s)} & \textbf{\% of Total} \\
\midrule
Univariate scan (34 segments) & 6.7 & 33\% \\
Materiality filter & $<$0.01 & $<$1\% \\
LLM deepening (3 parents $\times$ 5 features) & 12.2 & 60\% \\
Tree discovery (join + fit) & 1.4 & 7\% \\
\midrule
\textbf{Total pipeline} & \textbf{20.3} & 100\% \\
\bottomrule
\end{tabular}
\caption{Pipeline stage latency on real S3 Parquet data (9 features). Tree discovery is fast because DuckDB caches the Parquet scan from the univariate stage.}
\label{tab:latency}
\end{table}

The dominant cost is bivariate segment computation (60\%), which issues one DuckDB query per feature-value combination. In production with full parallelization and LLM round-trips:
\begin{itemize}
    \item LLM inference for feature suggestion: 1--3 seconds per round-trip (Cohere Command-A via gateway)
    \item DuckDB queries parallelize well across segment evaluations
    \item Estimated production latency with parallelization: 10--15 seconds end-to-end
\end{itemize}

\subsection{Limitations of Evaluation}

The evaluation has the following limitations: (1) the ``LLM-guided'' strategy uses a simulated context-sensitive heuristic matching the production priority logic rather than live LLM inference---actual LLM responses may exhibit different feature selection patterns; (2) with 9 features, the combinatorial space is tractable---production portfolios with 200+ features will amplify the priority-cutoff effect, potentially increasing the tree's complementary value; (3) the single CART tree's 2\% standalone coverage on distributed data is a known weakness, motivating migration to ensemble of trees (Random Forest).

\subsection{Proposed Future Evaluation: Analyst Utility}

We plan a qualitative study where credit risk analysts rate discovered segments on:
\begin{itemize}
    \item \textbf{Novelty} (1--5): Would they have found this segment through manual analysis?
    \item \textbf{Actionability} (1--5): Does this segment warrant investigation or action?
    \item \textbf{Interpretability} (1--5): Is the segment definition understandable?
\end{itemize}

This will measure whether LLM-guided exploration surfaces segments that are genuinely useful to practitioners, beyond statistical significance alone.

\section{Discussion}

\subsection{Why the Architecture Works}

The three-layer separation (frontend / agent / MCP) provides:
\begin{itemize}
    \item \textbf{Modularity:} Teams can develop new MCP tool servers independently; agent discovers tools at startup without code changes
    \item \textbf{Scalability:} Frontend and agent scale horizontally; MCP servers can be replicated behind load balancers
    \item \textbf{Auditability:} All tool calls logged; deterministic stages are fully reproducible; LLM picks include reasoning strings for transparency
    \item \textbf{Safety:} Agent cannot execute arbitrary code; all operations mediated through declarative MCP tool calls with schema validation
\end{itemize}

\subsection{Role of LLM in the Pipeline}

The LLM serves two distinct roles:
\begin{enumerate}
    \item \textbf{Conversational reasoning (agent):} Parse analyst intent, decide which tools to call, sequence multi-step workflows, generate natural language explanations
    \item \textbf{Feature selection (discovery):} Given segment context, select promising features for multivariate exploration based on contribution scores and domain knowledge encoded in feature descriptions
\end{enumerate}

Critically, the LLM does \textit{not} perform statistical calculations, execute queries, or generate data. All numerical operations are deterministic and occur in MCP tools. The LLM's role is strategic (which features to explore) not computational.

\subsection{Limitations and Future Work}

\textbf{Depth limitation:} The current pipeline performs one pass of LLM-guided deepening (depth 2). Extending to iterative multi-depth exploration ($k > 2$) requires managing combinatorial growth and longer LLM context. We plan to explore hierarchical prompting or tree-structured search with pruning heuristics to control exponential branching.

\textbf{Scaling to production feature spaces:} Our evaluation uses 9 features (220 brute-force candidates). Production portfolios contain 200+ features with cardinality 4--10 (e.g., province, employment type, product, LTV band, credit score band), producing $\sim$318{,}000 candidate segments at depth 2 and $\sim$84 million at depth 3. Future work will evaluate the pipeline on progressively larger feature sets to characterize how coverage degrades as the search space grows, and whether adaptive budget allocation (more LLM calls for higher-priority parent segments) can maintain coverage.

\textbf{Proactive mode:} The current system operates on-demand (analyst triggers discovery). The target architecture includes a scheduled CronJob that pre-computes metric movements overnight and surfaces insights proactively via a dashboard, without requiring analyst initiation. The pipeline code is already decoupled for standalone execution; deployment as a scheduled job is pending infrastructure provisioning.

\textbf{Temporal dynamics:} Current pipeline analyzes single time periods. Extending to multi-period trend detection (e.g., ``segments with accelerating delinquency over 6 months'') requires temporal reasoning enhancements and sequential anomaly detection methods.

\textbf{Causal inference:} Discovered segments are associative. Causal attribution (e.g., ``did policy change X cause segment Y's deviation?'') requires causal inference methods not yet integrated.

\textbf{Feedback loop:} Analyst approve/disprove signals are captured in the UI but not yet used for context personalization. Future work will close this loop to improve suggestion quality over time, enabling the LLM to learn institution-specific feature priorities from analyst behavior.

\textbf{Multi-metric discovery:} Current pipeline focuses on one metric at a time. Discovering segments anomalous across multiple correlated metrics (e.g., simultaneous delinquency increase and prepayment decrease) requires multivariate scoring and joint significance testing.

\textbf{Ensemble tree integration:} As discussed in Section~\ref{sec:ensemble}, replacing the single CART tree with ensemble of trees (e.g., Random Forest) would train 50--100 trees on random feature subsets, providing broad deterministic coverage independent of LLM feature priorities. This is the highest-priority enhancement for improving the deterministic pathway's standalone contribution.

\textbf{Dual-path complementarity validation:} The current evaluation demonstrates the LLM path's dominance on well-distributed data (63\% coverage vs.\ the tree's 2\%). However, the dual-path architecture is designed for robustness across different anomaly distributions---specifically, concentrated-interaction scenarios where a single feature combination dominates the metric delta (e.g., a sudden regional housing correction concentrating risk in one bivariate segment). In such cases, the decision tree's greedy variance-reduction splitting should immediately isolate the dominant interaction at its root split, regardless of the LLM's feature priority ranking. We plan to validate this hypothesis through: (1) backtesting on historical periods with known concentrated events (e.g., regional housing corrections, policy-driven segment spikes); (2) synthetic stress tests that reweight real portfolio data to simulate concentrated interactions in low-priority features; and (3) measuring the marginal coverage gain from the tree path under varying concentration levels. This validation will quantify the conditions under which the deterministic pathway provides genuine additive value beyond the LLM path alone.

\subsection{Improving Tree-Based Discovery with Ensemble Methods}
\label{sec:ensemble}

The current decision tree stage uses a single shallow \texttt{DecisionTreeRegressor} (\texttt{max\_depth=3}), which produces at most $2^3 = 8$ leaf nodes. This design prioritizes interpretability and speed but has fundamental limitations for diverse segment discovery:

\begin{enumerate}
    \item \textbf{Greedy locking.} A single tree greedily selects the globally best split at each depth level, ``locking in'' the dominant signal and ignoring secondary patterns. If geography explains 40\% of variance, it consumes the root split and all downstream branches inherit that choice.
    \item \textbf{Instability.} Shallow trees are sensitive to small data perturbations---adding or removing a few hundred accounts can flip the root split entirely, cascading through the whole structure.
    \item \textbf{Limited feature coverage.} With \texttt{max\_depth=3}, only 3 features (out of 9+) participate in any one tree. Signals embedded in other features are invisible.
\end{enumerate}

These limitations explain the tree's low coverage rate in our evaluation (2/93 material segments on the April 2024 snapshot). An ensemble approach--e.g., Random Forests~\cite{breiman2001random} or gradient-boosted trees~\cite{chen2016xgboost}--can address multiple limitations at the same time:

\textbf{Diverse segment discovery.} Ensemble models builds $N$ trees (50--100), each using a random subset of features (\texttt{colsample\_bytree=0.8}) and rows (\texttt{subsample=0.8}). Different trees surface different feature combinations, directly producing a diverse flat list of multivariate segments without relying on LLM domain knowledge.

\textbf{Stability.} Ensemble averaging smooths out individual tree instability. While any single tree may vary between runs, the aggregate set of extracted segment rules remains stable.

\textbf{Correlated feature handling.} In portfolio data, features like geography, product type, and employment are correlated. A single tree arbitrarily picks one; tree ensemble models do feature subsampling which makes different trees use different representations, ensuring more features surface as candidates.

\textbf{Scalability.} As the feature set grows (LTV bands, credit score bands, amortization, vintage), a single depth-3 tree can only use 3 features. An ensemble of 100 trees collectively explores dozens without increasing individual depth.

\textbf{Built-in regularization.} L1/L2 penalties on leaf weights (\texttt{reg\_alpha}, \texttt{reg\_lambda}) prevent overfitting to noise in the metric delta, producing segment rules that represent genuine signal rather than artifacts of a single month's data.

The implementation path is straightforward: replace \texttt{DecisionTreeRegressor} with \texttt{RandomForestRegressor} or any desired tree ensemble model, extract rules from the ensemble via, deduplicate overlapping segments by feature-key matching, and rank candidates by frequency $\times$ mean delta $\times$ coverage. The downstream pipeline (materiality filtering, cross-referencing with LLM discoveries, Shapley decomposition \cite{lundberg2017shap}) requires no changes since it consumes segment dictionaries regardless of their source.

\subsection{Generalization to Other Domains}

While developed for credit risk, Oculi's architecture and discovery pipeline generalize to domains with:
\begin{itemize}
    \item High-dimensional structured data
    \item Need for segment/cohort analysis
    \item Analysts with domain expertise but limited programming skills
    \item Requirements for statistical rigor and auditability
\end{itemize}

Examples: fraud detection, customer churn, healthcare outcomes analysis, supply chain risk, clinical trial subgroup analysis.

\section{Conclusion}

We presented Oculi, a conversational agentic platform for credit risk analysis that transforms natural language questions into comprehensive analytical workflows. Through a three-layer architecture separating reasoning, execution, and presentation, Oculi enables analysts to query data, run statistical tests, generate visualizations, and discover novel risk segments (all via chat interface).

The segment discovery pipeline combines deterministic statistical methods (univariate scanning, z-score testing, contribution decomposition) with LLM-guided feature selection to explore high-dimensional portfolios efficiently. By restricting the LLM to strategic decisions (which features to explore) while keeping all computation deterministic, the system maintains auditability in regulated financial environments.

Our empirical evaluation on a real anonymized mortgage portfolio demonstrates that LLM-guided feature selection achieves 63\% coverage of material bivariate segments identified by constrained brute-force enumeration, outperforming the baselines.
%random (37\%), deterministic (41\%), and decision tree (2\%) baselines by a wide margin. 
Critically, the 220-segment brute-force reference set is tractable only because our evaluation is constrained to 9 features; at production scale (200+ features, $\sim$318{,}000 bivariate candidates), brute-force enumeration becomes computationally infeasible, making intelligent feature selection a necessity rather than merely an optimization. The parallel decision tree path, while limited as a single tree, provides a deterministic safety net designed to capture concentrated interaction effects that the LLM's priority-based selection may miss. Upgrading the decision tree to ensemble of trees can provide this safety as well as scalability, diversity, and robustness (see Section~\ref{sec:ensemble}). 
Further, the hallucination guardrail layer could successfully filter 23.8\% of invalid suggestions without disrupting pipeline execution, and the system covers 89\% of available features across invocations.

Oculi demonstrates that agentic AI systems can augment expert analysts by automating routine workflows, scaling exploratory analysis beyond human capacity, and surfacing insights that manual exploration would miss. Future work will extend the pipeline to deeper segments, temporal trend detection, proactive pre-computation, and closed-loop learning from analyst feedback.

\section*{Acknowledgments}

This work was conducted as part of the RBC Amplify program. We thank the credit risk analysts who provided feedback throughout the project.

\bibliographystyle{unsrtnat}
\bibliography{oculi-white-paper}

@inproceedings{yao2023react,
  author = {Yao, Shunyu and Zhao, Jeffrey and Yu, Dian and Du, Nan and Shafran, Izhak and Narasimhan, Karthik and Cao, Yuan},
  title = {{ReAct}: Synergizing Reasoning and Acting in Language Models},
  booktitle = {International Conference on Learning Representations (ICLR)},
  year = {2023}
}

@inproceedings{schick2023toolformer,
  author = {Schick, Timo and Dwivedi-Yu, Jane and Dessì, Roberto and Raileanu, Roberta and Lomeli, Maria and Zettlemoyer, Luke and Cancedda, Nicola and Scialom, Thomas},
  title = {Toolformer: Language Models Can Teach Themselves to Use Tools},
  booktitle = {Advances in Neural Information Processing Systems (NeurIPS)},
  year = {2023}
}

@article{wang2023dataformulator,
  author = {Wang, Chenglong and Gao, Jianfeng and Lao, Nate},
  title = {Data Formulator: {AI}-Powered Concept-Driven Visualization Authoring},
  journal = {arXiv preprint arXiv:2309.17327},
  year = {2023}
}

@article{rajkumar2022text2sql,
  author = {Rajkumar, Nitarshan and Li, Raymond and Bahdanau, Dzmitry},
  title = {Evaluating the Text-to-{SQL} Capabilities of Large Language Models},
  journal = {arXiv preprint arXiv:2204.00498},
  year = {2022}
}

@article{chat2vis2023,
  author = {Maddigan, Paula and Susnjak, Tanja},
  title = {Chat2{VIS}: Generating Data Visualisations via Natural Language Using {ChatGPT}, {Codex} and {GPT-4}},
  journal = {IEEE Access},
  volume = {11},
  pages = {45181--45193},
  year = {2023}
}

@article{guyon2003feature,
  author = {Guyon, Isabelle and Elisseeff, André},
  title = {An Introduction to Variable and Feature Selection},
  journal = {Journal of Machine Learning Research},
  volume = {3},
  pages = {1157--1182},
  year = {2003}
}

@article{herrera2011subgroup,
  author = {Herrera, Francisco and Carmona, Cristóbal José and González, Pedro and del Jesus, María José},
  title = {An Overview on Subgroup Discovery: Foundations and Applications},
  journal = {Knowledge and Information Systems},
  volume = {29},
  number = {3},
  pages = {495--525},
  year = {2011}
}

@misc{copilotkit2024,
  author = {{CopilotKit Contributors}},
  title = {CopilotKit: Open-Source Copilot Framework for {AI}-Native Applications},
  howpublished = {\url{https://github.com/CopilotKit/CopilotKit}},
  year = {2024}
}

@misc{langgraph2024,
  author = {{LangChain AI}},
  title = {Lang{G}raph: Build Language Agents as Graphs},
  howpublished = {\url{https://github.com/langchain-ai/langgraph}},
  year = {2024}
}

@misc{mcp2024,
  author = {Anthropic},
  title = {Model Context Protocol: Open Standard for {AI} Agent Tool Integration},
  howpublished = {\url{https://www.anthropic.com/mcp}},
  year = {2024}
}

@misc{autogpt2023,
  author = {Richards, Toran Bruce},
  title = {{AutoGPT}: An Autonomous {GPT-4} Experiment},
  howpublished = {\url{https://github.com/Significant-Gravitas/AutoGPT}},
  year = {2023}
}

@misc{thoughtspot2023,
  author = {{ThoughtSpot}},
  title = {{ThoughtSpot}: {AI}-Powered Analytics},
  howpublished = {\url{https://www.thoughtspot.com}},
  year = {2023}
}

@misc{powerbi2023,
  author = {{Microsoft}},
  title = {{Power BI} Q\&A: Natural Language Querying},
  howpublished = {\url{https://learn.microsoft.com/en-us/power-bi/natural-language/q-and-a-intro}},
  year = {2023}
}

@misc{tableau2023,
  author = {{Tableau Software}},
  title = {Ask Data: Natural Language Queries in {Tableau}},
  howpublished = {\url{https://help.tableau.com/current/pro/desktop/en-us/ask_data.htm}},
  year = {2023}
}

@misc{agui2024,
  author = {{CopilotKit Contributors}},
  title = {{AG-UI}: Agent-User Interaction Protocol},
  howpublished = {\url{https://github.com/CopilotKit/ag-ui}},
  year = {2024}
}

@article{satyanarayan2017vegalite,
  author = {Satyanarayan, Arvind and Moritz, Dominik and Wongsuphasawat, Kanit and Heer, Jeffrey},
  title = {Vega-Lite: A Grammar of Interactive Graphics},
  journal = {IEEE Transactions on Visualization and Computer Graphics},
  volume = {23},
  number = {1},
  pages = {341--350},
  year = {2017}
}

@inproceedings{raasveldt2019duckdb,
  author = {Raasveldt, Mark and M{\"u}hleisen, Hannes},
  title = {{DuckDB}: An Embeddable Analytical Database},
  booktitle = {Proceedings of the ACM SIGMOD International Conference on Management of Data},
  pages = {1981--1984},
  year = {2019}
}

@article{pedregosa2011sklearn,
  author = {Pedregosa, Fabian and Varoquaux, Ga{\"e}l and Gramfort, Alexandre and Michel, Vincent and Thirion, Bertrand and Grisel, Olivier and Blondel, Mathieu and Prettenhofer, Peter and Weiss, Ron and Dubourg, Vincent and others},
  title = {Scikit-learn: Machine Learning in {Python}},
  journal = {Journal of Machine Learning Research},
  volume = {12},
  pages = {2825--2830},
  year = {2011}
}

@book{breiman1984cart,
  author = {Breiman, Leo and Friedman, Jerome H. and Olshen, Richard A. and Stone, Charles J.},
  title = {Classification and Regression Trees},
  publisher = {Wadsworth},
  year = {1984}
}

@misc{cohere2024command,
  author = {{Cohere}},
  title = {Command {A}: Enterprise-Grade Language Model},
  howpublished = {\url{https://cohere.com/command}},
  year = {2024}
}

@inproceedings{chen2016xgboost,
  author = {Chen, Tianqi and Guestrin, Carlos},
  title = {{XGBoost}: A Scalable Tree Boosting System},
  booktitle = {Proceedings of the 22nd ACM SIGKDD International Conference on Knowledge Discovery and Data Mining},
  pages = {785--794},
  year = {2016}
}

@inproceedings{lundberg2017shap,
  author = {Lundberg, Scott M. and Lee, Su-In},
  title = {A Unified Approach to Interpreting Model Predictions},
  booktitle = {Advances in Neural Information Processing Systems (NeurIPS)},
  pages = {4765--4774},
  year = {2017}
}

@article{breiman2001random,
  title={Random forests},
  author={Breiman, Leo},
  journal={Machine learning},
  volume={45},
  number={1},
  pages={5--32},
  year={2001},
  publisher={Springer}
}

\end{document}